\documentclass[conference]{IEEEtran}
\IEEEoverridecommandlockouts
\usepackage{cite}
\usepackage{amsmath,amssymb,amsfonts}
\usepackage{algorithmic}
\usepackage{graphicx}
\usepackage{booktabs}
\usepackage[caption=false,font=footnotesize]{subfig}
\usepackage{textcomp}
\usepackage{xcolor}
\def\BibTeX{{\rm B\kern-.05em{\sc i\kern-.025em b}\kern-.08em
    T\kern-.1667em\lower.7ex\hbox{E}\kern-.125emX}}
\begin{document}

\title{Residual Modeling for High-Fidelity Learned Compression of Scientific Data\\

}

\author{
\IEEEauthorblockN{Liangji Zhu, Sanjay Ranka, and Anand Rangarajan}
\IEEEauthorblockA{
\textit{Department of Computer \& Information Science \& Engineering}\\
\textit{University of Florida}\\
Gainesville, FL, USA\\
zhu.liangji@ufl.edu, ranka@cise.ufl.edu, anand@cise.ufl.edu
}
}

\maketitle

\begin{abstract}
Lossy compression is essential for the massive spatiotemporal data produced by modern scientific simulations. Machine-learning-based compressors achieve high compression ratios at moderate accuracy targets, but their training objectives minimize aggregate reconstruction error and provide no guarantee on any individual block. Scientific applications require per-block accuracy targets, so a deployable learned compressor must add a per-block residual correction stream. Guaranteed Autoencoder (GAE) techniques close this gap by retaining the smallest number of SVD/PCA-style coefficients per block needed to meet the target. This works well at moderate tolerances. In the high-fidelity regime, with block-level NRMSE between $10^{-6}$ and $10^{-4}$, however, the number of retained GAE coefficients per block grows rapidly, the correction stream dominates the total rate, and the advantage of the learned base model erodes.

This paper proposes a different view: the learned residual is a \emph{structurally distinct object} from the original scientific field, with smaller dynamic range and different local spatiotemporal statistics, and should be coded by a representation designed for the learned residual rather than by a global per-block correction basis. We instantiate this view with two complementary residual coders. 
\emph{LBRC (Lorenzo-Based Residual Coding)} is a deterministic, training-free pipeline that adaptively quantizes the learned residual to a target block-level NRMSE and losslessly encodes the resulting integer residual using 3D Lorenzo differencing, zigzag mapping, bit-plane coding, and entropy coding. \emph{NGLR (Neural-Guided Lorenzo Residual Coding)} adds a causal neural predictor that outputs a normalized bias, which is rescaled and used inside an integer-rounded Lorenzo prediction in the same deterministic integer pipeline, reducing the entropy of the remaining residual code stream while preserving deterministic decoding of the quantized residual.
The resulting design combines target-matched integer Lorenzo coding with a collection-adaptive neural bias predictor whose weights are charged to the bitstream.

Across E3SM, JHTDB, and ERA5 datasets at block-level NRMSE targets from $10^{-6}$ to $10^{-4}$, LBRC improves compression ratio over GAE by 30--60\% and is broadly competitive with SZ; NGLR adds a further 10--40\% over LBRC and outperforms SZ in the evaluated high-fidelity regime. Residual representations tailored to learned-compressor residuals preserve the advantage of learned compression where global residual correction would otherwise become rate-dominant.
\end{abstract}


\section{Introduction}
\label{sec:intro}
Large-scale scientific simulations in domains such as climate modeling~\cite{e3sm,era5}, fluid dynamics~\cite{jhtdb}, and combustion~\cite{s3d} routinely produce terabytes to petabytes of data per run. The growing gap between data production rates and the I/O and storage bandwidth of modern high-performance computing (HPC) systems makes lossy compression an essential component of scientific workflows~\cite{cappello2019use}. Unlike image and video compression for human consumption, scientific compression must provide reliable accuracy guarantees so that the reconstructed data can be used in downstream simulations and analyses without compromising scientific conclusions.

Compression approaches for scientific data fall into two broad families. Traditional compressors such as SZ~\cite{SZ_3,sz3}, ZFP~\cite{zfp,fox2020stability}, MGARD~\cite{MGARD_2,gong2023mgard}, and FAZ~\cite{liu2023faz} enforce a pointwise or norm-based error bound on every block, but their local or blockwise models do not exploit cross-block learnable structure, so their compression ratios can plateau. Learned compressors~\cite{liu2021high,minnen2018joint,vae_z,xiao_bigdata,li2025foundation,caesar,li2025generative} achieve higher compression ratio (CR) at moderate accuracy targets by learning compact latent representations, but their training objectives minimize an aggregate, \emph{macro} loss and do not, by themselves, guarantee any individual block. Scientific applications require \emph{per-block} accuracy guarantees, because downstream analyses can be locally sensitive to a single outlier block.

Closing this gap requires a per-block residual correction stream that, when added to the learned reconstruction $\tilde{x}$, brings the block's reconstruction error below the user-specified target. The Guaranteed Autoencoder (GAE) framework~\cite{jaemoon3} introduced an elegant route to this guarantee: the decoder of an autoencoder with piecewise-linear units can be represented exactly as a block-specific linear operator $L_i$, and the block residual $u_i = x_i - \tilde{x}_i$ can be projected onto the SVD basis of $L_i$. Coefficients of this projection are then retained in descending order of magnitude until the residual reconstructed from the retained coefficients brings the block error below the target. This procedure---per-block SVD projection plus adaptive coefficient retention---provides the block-level accuracy guarantee used both in the original GAE work and in the more recent CAESAR framework~\cite{caesar}.

The GAE-style correction works very well when only a few coefficients per block are needed, i.e., in the moderate-to-high compression regime (block-level NRMSE $\geq 10^{-3}$). In this regime, the per-block correction cost is small, the learned base compressor dominates the rate budget, and the combined system outperforms traditional compressors such as SZ by 2--10$\times$~\cite{caesar}. As the target tightens into the high-fidelity, low-compression regime (block-level NRMSE between $10^{-6}$ and $10^{-4}$, increasingly demanded by domain scientists~\cite{gong2023mgard,cappello2019use}), the number of basis coefficients per block grows rapidly, the per-block correction stream dominates the total rate, the advantage of the learned base model erodes, and the overall CR can fall below that of SZ. The bottleneck shifts from the base reconstruction to the per-block correction itself: what is needed is a more compact per-block residual representation, not a stronger base model. This is not a weakness of GAE-style correction in general: prior GAE/CAESAR results show that learned compression is highly effective at moderate targets (around $10^{-3}$ and looser), where the base model dominates and only a small residual correction is needed. The limitation addressed here is specific to the stricter low-NRMSE regime, where global SVD/PCA-style correction becomes rate-limiting.

The block residual $u_i$ is not random noise: it inherits local smoothness, anisotropic gradients, and spatiotemporal correlations from the underlying field. Treating it as an unstructured vector and using a global per-block basis is wasteful; a representation that explicitly models local prediction along the temporal and spatial axes can use far fewer bits while still meeting the same per-block accuracy target. This suggests a hybrid predictive design, similar in spirit to LPCNet~\cite{valin2019lpcnet}, where a fixed linear predictor removes the easily predictable component and a neural model focuses on the remaining residual.

To this end, rather than forcing a single end-to-end neural network to balance a delicate Pareto tradeoff between rate and distortion, our framework enforces a strict separation of concerns. We first quantize the residual between the base reconstruction and the true data such that blockwise error bounds are strictly and deterministically satisfied. We then losslessly encode the resulting integer residual field using a classical linear predictor. Because linear stencils leave structured, non-linear innovation footprints, we introduce a lightweight neural network to dynamically model the Lorenzo prediction bias using a localized, causal decoding context. In this design, the neural network acts purely as an entropy reducer—assisting in the lossless compression of the residual stream without ever endangering the underlying physical guarantee. 

Lorenzo-based residual correction (LBRC) isolates the deterministic component of this idea by applying target-matched residual quantization, 3D Lorenzo differencing, bit-plane coding, and entropy coding directly to the learned residual. We further hypothesize that the Lorenzo prediction error itself remains partially predictable from the base reconstruction and causal residual context. Neural-guided Lorenzo Residual Coding (NGLR) is driven by this intuition: it extends LBRC with a neural Lorenzo bias predictor that targets this remaining predictability to reduce the entropy of the residual code stream while preserving deterministic decoding.

In sum, we propose two complementary per-block residual representations that can attach as a post-processing module to learned block compressors; we instantiate the approach with CAESAR-V~\cite{caesar} and replace the GAE-style SVD-basis projection. Our contributions are:

\begin{itemize}
\item \textbf{A residual-centric view of high-fidelity learned compression.} We show that in the low-NRMSE regime the rate bottleneck has shifted from the learned base reconstruction to the per-block residual correction stream, so that further progress requires a better residual representation rather than a stronger base model.

\item \textbf{LBRC: deterministic coding of learned residuals.} We introduce a training-free residual representation that adaptively quantizes the learned residual to match a target NRMSE and applies 3D Lorenzo differencing, zigzag mapping, bit-plane coding, and entropy coding. LBRC is both a practical method and the deterministic ablation of the proposed residual representation.

\item \textbf{NGLR: neural-guided residual prediction.} We introduce a causal neural bias predictor that uses base-reconstruction features and already-decoded residual context to predict a normalized bias for an integer-rounded
Lorenzo prediction, reducing the entropy of the residual code stream while preserving deterministic decoding.
\end{itemize}

We evaluate LBRC and NGLR on three large-scale scientific datasets---E3SM (climate)~\cite{e3sm}, JHTDB (turbulence)~\cite{jhtdb}, and ERA5 (atmospheric reanalysis)~\cite{era5}---across NRMSE targets from $10^{-6}$ to $10^{-4}$. LBRC improves CR over GAE by 30--60\% and is broadly competitive with SZ; NGLR adds a further 10--40\% over LBRC and outperforms SZ in the evaluated high-fidelity regime. The resulting design combines target-matched integer Lorenzo coding with a collection-adaptive neural bias predictor whose weights are charged to the bitstream.

The remainder of the paper is organized as follows. Section~\ref{sec:related} surveys related work; Section~\ref{sec:method} presents LBRC and NGLR; Section~\ref{sec:experiments} reports experimental results; and Section~\ref{sec:conclusion} concludes and discusses future work.

\section{Related Work}
\label{sec:related}

\subsection{Error-Bounded Lossy Compression of Scientific Data}
Domain scientists require strict bounds on reconstruction error so that compressed data can be reused for downstream analyses without altering scientific conclusions~\cite{cappello2019use}. Three classical families dominate the field: prediction-based (SZ~\cite{SZ_3,sz3}, often using a Lorenzo predictor~\cite{Lorenzo}), block-transform (ZFP~\cite{zfp,fox2020stability}), and multilevel (MGARD~\cite{MGARD_2,gong2023mgard}); FAZ~\cite{liu2023faz} is a modular auto-tuned framework combining predictive and wavelet components. These methods offer rigorous guarantees and low overhead, but their compression ratios can plateau when cross-block structure is available but not explicitly modeled.

\subsection{Learned and Generative Models for Scientific Compression}
Learned image codecs with quantization-aware training and hyperprior models~\cite{minnen2018joint,vae_z} have been adapted to scientific data via fully connected~\cite{liu2021high} and convolutional~\cite{Hayne9671627} autoencoders and attention-based block coders~\cite{xiao_bigdata}. To meet strict error bounds, hybrid pipelines such as AE-SZ~\cite{ae-sz} and AETMC~\cite{Jaemoon} couple an autoencoder with SZ or MGARD as a residual coder. The Guaranteed Autoencoder (GAE)~\cite{jaemoon3} provides block-level error bounds by representing the decoder as a block-specific linear operator and projecting the residual onto the SVD basis of that operator. CAESAR~\cite{caesar} combines a variational autoencoder with hyperpriors (CAESAR-V) and a latent diffusion model (CAESAR-D), with a GPU-parallel post-processing module that enforces error bounds; it achieves up to 10$\times$ higher CR than SZ3 at moderate targets. As the target NRMSE tightens, however, the PCA-style correction module  dominates the rate budget, motivating this  work.

\subsection{Residual Coding and Lorenzo-Based Prediction}
Residual coding has a long history in image, video, and scientific compression. In SZ, the Lorenzo predictor~\cite{Lorenzo} estimates the value at a grid point from its already-decoded $\ell_1$-neighbors and encodes the integer-quantized prediction error. Bit-plane coding has been used successfully in ZFP, JPEG2000, and MGARD-style frameworks to compress integer residuals one significance level at a time. For learned compressors, the dominant residual coding approach has been a global linear projection onto a basis derived from the decoder Jacobian, as in GAE~\cite{jaemoon3} and the CAESAR post-processing module~\cite{caesar}. This global formulation is elegant and provides block-level guarantees, but the cost per block scales with the number of coefficients required to meet the error bound, which can be large in the high-fidelity regime.

Our work takes a different view of these prior approaches. Unlike traditional compressors such as SZ, ZFP, and MGARD, which operate directly on the original scientific field, our methods operate on the \emph{learned residual} after a base reconstruction has already removed the dominant large-scale structure. This learned residual has smaller dynamic range and different local statistics than the original field. Our results show that a residual representation designed for this field can use significantly fewer bits than a global per-block PCA basis, and can make the overall learned-compression pipeline competitive with or better than SZ in the high-fidelity regime. LBRC therefore uses 3D Lorenzo differencing as a residual representation for the learned residual rather than as a full-field predictor. NGLR further conditions the prediction on base-reconstruction features and causal residual context, exploiting structure in the learned residual that a fixed Lorenzo stencil cannot capture, while keeping the encoded code stream deterministic and exactly decodable. The causal $q$-neighbor context branch in NGLR is structurally related to autoregressive context models in learned image compression~\cite{van2016conditional,minnen2018joint}, but it operates on the integer-quantized residual rather than continuous latents and is integrated with a deterministic bit-plane coder, so the entire pipeline remains exactly decodable under a block-level NRMSE target.

\section{Methodology}
\label{sec:method}

\subsection{Problem Formulation}
\label{sec:formulation}
Let $x \in \mathbb{R}^{T \times H \times W}$ denote a spatiotemporal scientific data block with $T$ temporal frames and spatial resolution $H \times W$ (extensions to additional channels and 3D spatial volumes are straightforward). Let $\mathcal{C}$ denote a base learned compressor with encoder $\mathcal{E}$ and decoder $\mathcal{D}$, producing a base reconstruction
\begin{equation}
\tilde{x} = \mathcal{D}(\mathcal{Q}_{\mathrm{lat}}(\mathcal{E}(x))),
\end{equation}
where $\mathcal{Q}_{\mathrm{lat}}$ denotes the latent-space quantization. For a user-specified normalized error target $\tau$, the goal of residual coding is to produce a side stream that, when combined with $\tilde{x}$, yields a final reconstruction $\hat{x}$ satisfying
\begin{equation}
\frac{\|x - \hat{x}\|_2}{\sqrt{N} \cdot (x_{\max} - x_{\min})} \leq \tau,
\label{eq:nrmse}
\end{equation}
where $N$ is the number of voxels and the denominator is the data range used to normalize the error. We refer to the quantity on the left of \eqref{eq:nrmse} as the NRMSE. The residual is
\begin{equation}
u = x - \tilde{x}.
\end{equation}
In the high-fidelity regime ($\tau$ between $10^{-6}$ and $10^{-4}$), $u$ is small in magnitude but still contains significant local structure inherited from the underlying physics. Our methods encode a quantized version of \(u\). At decode time, the correction stream recovers an integer residual field \(\hat{q}\), which is dequantized as \(\Delta \hat{q}\) and added back to the base reconstruction:
\begin{equation}
\hat{x} = \tilde{x} + \Delta \textcolor{blue}{\cdot} \hat{q}.
\end{equation}
The quantization step \(\Delta\) is selected so that the final reconstruction satisfies the block-level NRMSE target in (2).

\subsection{LBRC: Training-Free 3D Lorenzo Residual Coding}
\label{sec:lbrc}
\begin{figure*}[t]
\centering
\includegraphics[width=0.7\textwidth]{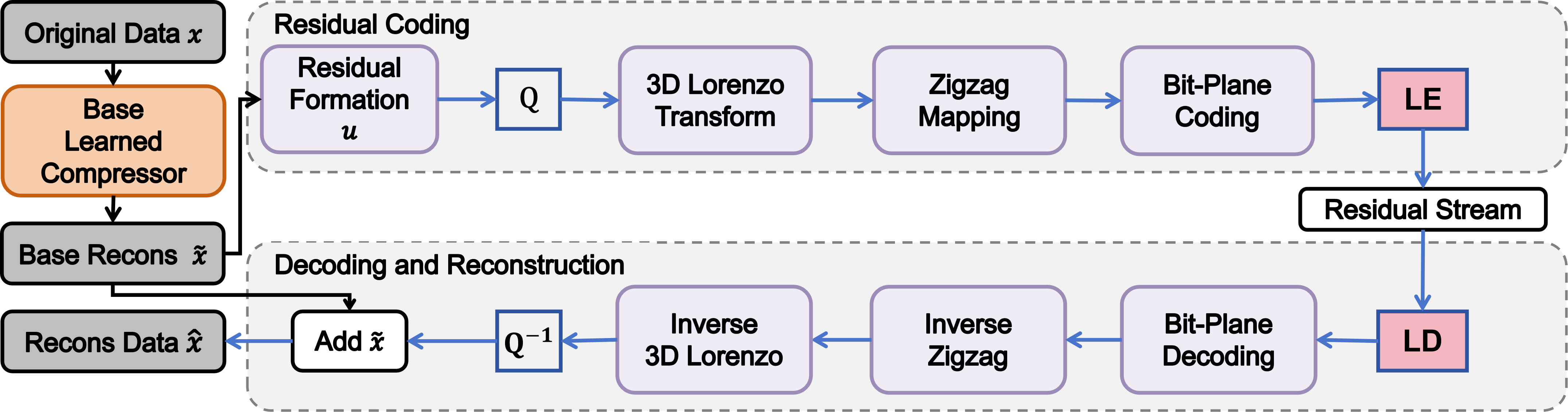}
\caption{Overview of the Lorenzo-Based Residual Coding (LBRC) pipeline. The base learned compressor produces $\tilde{x}$, and the residual $u = x - \tilde{x}$ is quantized (Q), transformed by a 3D Lorenzo operator, zigzag-mapped, bit-plane coded, and entropy-encoded (LE). The decode path (LD$\rightarrow$Add $\tilde{x}$) is the exact inverse.}
\vspace*{-0.6cm}
\label{fig:lbrc}
\end{figure*}

LBRC is a deterministic, training-free pipeline composed of five steps---adaptive quantization, 3D Lorenzo transform, zigzag mapping, bit-plane coding, and lossless entropy coding---followed by their exact inverses on the decode side. Figure~\ref{fig:lbrc} illustrates the full pipeline. Together, these steps realize the target-matched integer Lorenzo coder referenced in the abstract: integer-domain throughout, exactly invertible, and matched to a user-specified block-level NRMSE.

\subsubsection{Adaptive Residual Quantization}
Given the residual $u$ and a target NRMSE $\tau$, we quantize $u$ uniformly with step size $\Delta$:
\begin{equation}
\hat{u}_i = \mathrm{round}(u_i / \Delta) \cdot \Delta.
\end{equation}

We choose the largest $\Delta$ by one-dimensional binary search: for each candidate, we quantize, reconstruct $\hat{x}=\tilde{x}+\Delta q$, and accept the largest $\Delta$ whose achieved block-level NRMSE is no larger than $\tau$. This target-matched selection avoids overly conservative compression where the achieved error is far below the requested target.

After quantization, we work with the integer residual field $q = \hat{u}/\Delta \in \mathbb{Z}^{T \times H \times W}$.

\subsubsection{3D Lorenzo Transform}
The Lorenzo predictor~\cite{Lorenzo}, widely used in SZ, predicts a grid value from its already-decoded $\ell_1$-neighbors. In the 3D case, the predicted value at position $(t,h,w)$ is
\begin{align}
L_{t,h,w} =\ & q_{t-1,h,w} + q_{t,h-1,w} + q_{t,h,w-1} \nonumber \\
& - q_{t-1,h-1,w} - q_{t-1,h,w-1} - q_{t,h-1,w-1} \nonumber \\
& + q_{t-1,h-1,w-1}.
\end{align}
The Lorenzo-transformed residual $d_{t,h,w} = q_{t,h,w} - L_{t,h,w}$ is then stored in place of $q$. Because Lorenzo is an exact integer operator with a fixed causal stencil, applying it to the integer residual produces another integer field, fully invertible without rounding error.

Applying Lorenzo to a quantized residual field rather than directly to the data (as in SZ) is effective because $u$ already has small magnitude and concentrated structure after subtraction of $\tilde{x}$, and neighboring residuals share signs and directional trends inherited from the underlying field. The Lorenzo difference $d$ is therefore tightly peaked around zero, which the subsequent bit-plane and entropy coding stages exploit.

\subsubsection{Zigzag Mapping and Bit-Plane Coding}
The signed Lorenzo differences $d$ are mapped to non-negative integers via the zigzag bijection
\begin{equation}
z(d) = \begin{cases} 2d, & d \geq 0 \\ -2d - 1, & d < 0 \end{cases}
\end{equation}
which preserves the small-magnitude concentration. The non-negative field $z(d)$ is decomposed into bit planes; because the distribution of $d$ is sharply concentrated near zero, the high-order planes are sparse. Bit-plane streams are concatenated and passed to a lossless entropy coder (LE in Fig.~\ref{fig:lbrc}).

\subsubsection{Decoding}
The decoder reverses each step exactly: Lossless decoding (LD), inverse bit-plane reconstruction, inverse zigzag, inverse 3D Lorenzo, dequantization ($\mathcal{Q}^{-1}$), and addition to the base reconstruction $\tilde{x}$. Since each step is integer-deterministic and invertible, LBRC introduces no extra error beyond the quantization step. The final reconstruction $\hat{x}$ satisfies the NRMSE target by construction of $\Delta$.

\textbf{Guarantee.} For a fixed base reconstruction \(\tilde{x}\), LBRC selects the largest quantization step \(\Delta\) for which the corrected block \(\hat{x}=\tilde{x}+\Delta q\) satisfies the requested NRMSE target \(\tau\) in the sense of \eqref{eq:nrmse}. Since all subsequent coding stages are exactly invertible over \(q\), the decoded reconstruction satisfies the same block-level NRMSE target selected during encoding. We emphasize that this guarantee is on the block-level NRMSE, not on pointwise error, which is the standard guarantee provided by SZ-family compressors; the two are related but not equivalent (see Section~\ref{sec:experiments}).

LBRC is attractive as a baseline: it has no training cost, no learned parameters, deterministic behavior across runs, and a small, predictable runtime. 
As we show in Section~\ref{sec:experiments}, it already achieves compression ratios competitive with SZ in the high-fidelity regime, and better than GAE-style global PCA correction.

\subsection{NGLR: Neural-Guided Lorenzo Residual Coding}
\label{sec:nglr}

NGLR is motivated by the observation that the Lorenzo residual code is still partially predictable. The base codec first removes the dominant large-scale structure, and a fixed Lorenzo stencil further reduces the learned residual, but the remaining prediction errors are not always random. They can contain systematic patterns related to the base reconstruction and the causal residual context. NGLR therefore keeps the deterministic LBRC backend and learns only a lightweight causal bias to correct these systematic Lorenzo errors, making the remaining integer code smaller while preserving deterministic decoding. The network output is continuous and normalized; it is rescaled
to the residual-code domain, and the guided Lorenzo prediction is rounded to an integer before the remaining code is stored. Thus, the entire NGLR pipeline remains deterministic; and the bias predictor is \emph{collection-adaptive}---trained on the data being compressed and charged to the bitstream as part of the compressed representation.
%

For voxel \(i\), the network outputs a normalized bias
\begin{equation}
\bar{b}_i=f_{\theta}(\phi(\tilde{x})_i,c_i),
\end{equation}
where \(\phi(\tilde{x})_i\) denotes features from the base reconstruction and \(c_i\) denotes causal residual-context features.
The network output $\bar{b}_i$ is a normalized floating-point bias; it is rescaled by a separate residual-code scale $s_d$ (defined in the training objective below) as
\begin{equation}
b_i = s_d \, \bar{b}_i.
\end{equation}
This scale $s_d$ is not the residual quantization step $\Delta$: $\Delta$ maps the continuous residual $u$ to the integer code $q$, whereas $s_d$ only converts the normalized neural bias to the integer-code scale. Given the integer Lorenzo estimate $L_i$, NGLR forms
\begin{equation}
p_i=\mathrm{round}(L_i+b_i), \qquad \delta_i=q_i-p_i.
\end{equation}
Only the integer code \(\delta_i\) is stored. The same $s_d$ is used at training, encoding, and decoding, so the rescaling is a fixed operation rather than a learned one.

\begin{figure*}[t]
\centering
\includegraphics[width=0.7\textwidth]{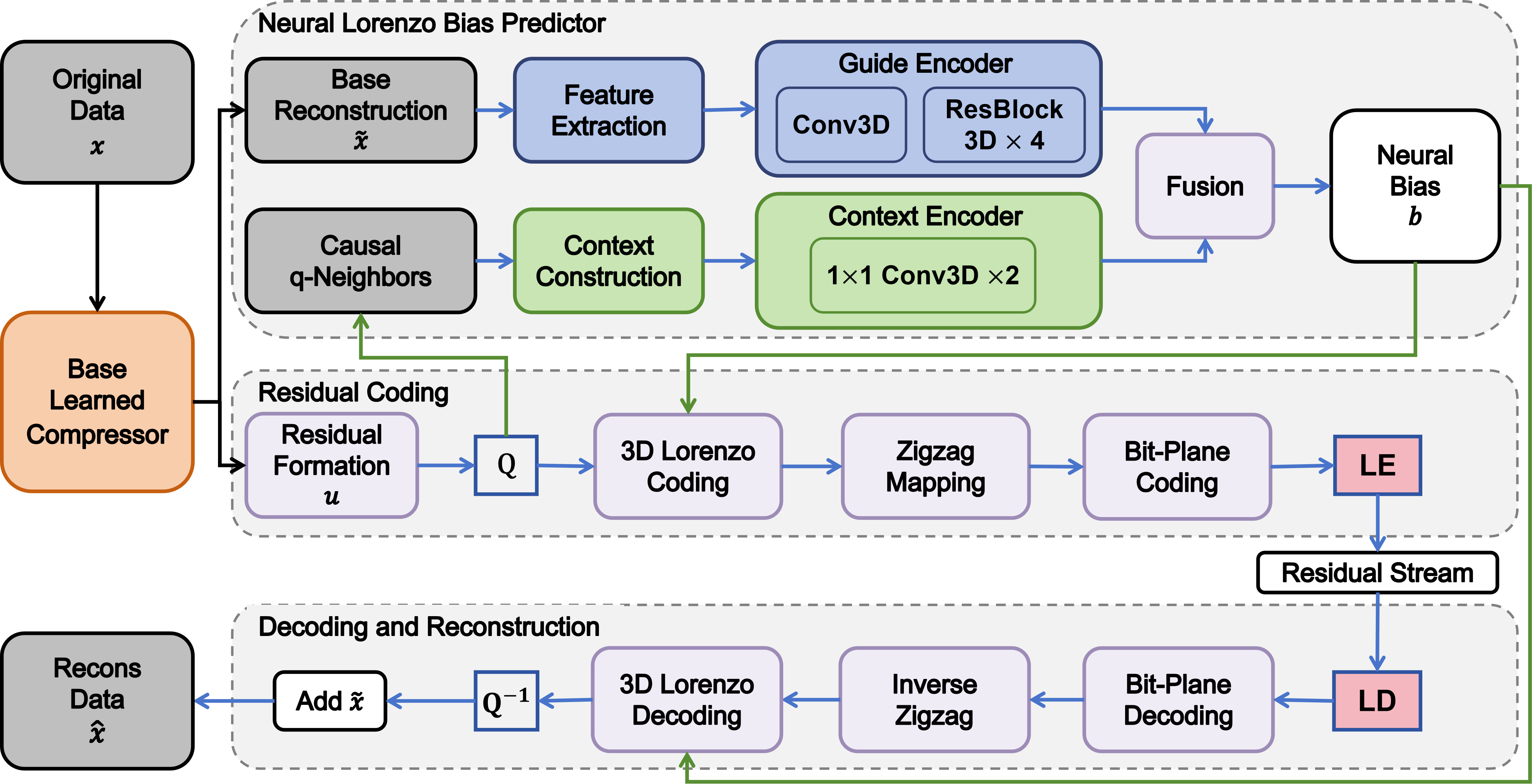}
\caption{
Overview of Neural-Guided Lorenzo Residual Coding (NGLR). The Neural Lorenzo Bias Predictor fuses Guide features from the base reconstruction \(\tilde{x}\) and causal \(q\)-neighbor context to produce a per-voxel bias \(b\). The bias refines the Lorenzo estimate, and only the remaining integer code \(\delta=q-\mathrm{round}(L+b)\) is stored. The same causal context is reconstructed during decoding from already decoded neighbors.
}
\label{fig:nglr}
\vspace*{-0.6cm}
\end{figure*}

Figure~\ref{fig:nglr} shows the full NGLR architecture. Compared with LBRC, NGLR keeps the same residual quantization and integer-code backend, but uses a \emph{Neural Lorenzo Bias Predictor} to guide the Lorenzo prediction before the remaining code is stored. 
The same causal context is reconstructed during decoding, making the bias reproducible.

\subsubsection{Information Branches}

The two information branches provide complementary signals: the reconstruction branch supplies continuous-field structure from the decoded approximation, and the causal $q$-neighbor branch supplies code-domain context from already decoded quantized residuals. Combining them lets NGLR correct systematic biases in the fixed Lorenzo prediction using both the structure of the reconstructed field and the local pattern of the integer code, concentrating the remaining stored correction near zero.

\textbf{(i) Guided branch from $\tilde{x}$.} A feature extractor followed by a Guide Encoder (a Conv3D layer plus four 3D residual blocks) maps the base reconstruction $\tilde{x}$ to a feature volume that summarizes local field structure where the residual is being encoded. Since $\tilde{x}$ is produced by the base decoder on both the encoder and decoder sides, this branch does not require an additional side stream.

\textbf{(ii) Causal context branch from $q$-neighbors.} We construct a causal residual context $c_i$ containing the seven first-order causal $q$-neighbors and the corresponding standard Lorenzo estimate $L_i$. A lightweight Context Encoder maps this context to a feature vector at each voxel. Because the context uses only causally preceding neighbors, the same information is available during decoding and no future residuals  are leaked.

\subsubsection{Fusion and Bias Generation}

After the Guide Encoder and the Context Encoder produce their feature representations, the Fusion module combines them at each voxel to generate the neural bias. Let \(g_i\) denote the Guide feature extracted from the base reconstruction, and let \(h_i\) denote the encoded causal-context feature. The two features are concatenated along the channel dimension as \(m_i=[g_i,h_i]\). A lightweight pointwise network, implemented with \(1\times1\times1\) Conv3D layers and nonlinear activations, maps \(m_i\) to a scalar normalized bias $\bar{b}_i$. Since this fusion is pointwise, it only mixes the two feature streams at the same voxel; all causal information is determined by the context construction. The resulting $\bar{b}_i$ is rescaled to the integer-residual domain by $b_i = s_d \bar{b}_i$ and used as the neural correction to the Lorenzo estimate.

\subsubsection{Encoding and Decoding Workflow}

At encode time, NGLR follows a fixed causal wavefront scan order. (1) Compute Guide features from the baseline reconstruction \(\tilde{x}\). (2) For each voxel, construct the causal context from the seven available \(q\)-neighbors and the standard Lorenzo estimate. (3) The neural bias predictor outputs $\bar{b}_i$, which is rescaled to $b_i = s_d \bar{b}_i$ and used to form the guided prediction \(p_i=\mathrm{round}(L_i+b_i)\). (4) Store only the remaining integer code \(\delta_i=q_i-p_i\), using the same zigzag, bit-plane, and lossless coding backend as LBRC.

At decode time, the same scan order is used. (1) Decode the next stored code \(\delta_i\). (2) Reconstruct the same causal context from already decoded \(\hat{q}\)-neighbors and compute the same neural bias from \(\tilde{x}\). (3) Reproduce the guided prediction \(\hat{p}_i=\mathrm{round}(L_i+b_i)\). (4) Recover \(\hat{q}_i=\hat{p}_i+\delta_i\), insert it into the decoded residual field, and continue.

In our implementation, deterministic decoding is enforced by fixed weights, deterministic evaluation mode, fixed numerical precision, a fixed residual-code scale $s_d$, and a fixed rounding rule (we round half to even). The base decoder producing $\tilde{x}$ is also run deterministically. Under these conditions both sides compute the same $p_i$, and the stored integer code $\delta_i$ exactly recovers $q_i$.

\subsubsection{Training Objective}

The neural bias predictor is trained to reduce the magnitude of the remaining predictive code. Let
\begin{equation}
d_i = q_i - L_i
\end{equation}
be the standard Lorenzo residual code at voxel \(i\). The network predicts a normalized bias \(\bar{b}_i\) to approximate \(d_i/s_d\), where $s_d$ is the standard deviation of the Lorenzo residual codes on the training blocks; it is computed once per compressed collection and stored as scalar metadata. We optimize a Charbonnier loss:
\begin{equation}
\mathcal{L}
=
\frac{1}{N}\sum_i\sqrt{\left(\frac{d_i}{s_d} - \bar{b}_i\right)^2 + \epsilon^2
}.
\end{equation}
This loss encourages the neural bias to explain predictable structure in the Lorenzo residual code. Although it is not a direct entropy objective, reducing the magnitude of the remaining integer code typically leads to sparser high-order bit planes and a smaller correction stream.

The default NGLR predictor uses \(32\) reconstruction hidden channels, \(16\) q-context hidden channels, and four 3D residual blocks. This gives approximately \(2.3\times 10^5\) trainable parameters. The serialized model size is about \(0.94\) MB and is included in the total compressed size.
\subsubsection{Cost and Practical Considerations}
\label{sec:nglr-cost}
The compressed representation of NGLR consists of the base latent stream, the residual correction stream, and the serialized neural bias predictor. The neural model is counted as part of the bitstream rather than treated as a free external prior. This makes NGLR a \emph{collection-adaptive} residual coder: the bias predictor is learned from the data being compressed, serialized once, and charged to the compressed representation. 
This is analogous to transmitting a collection-specific codebook or model: adaptation is allowed because the learned parameters are serialized and included in the compressed representation. 
We note that compression ratio counts stored bits, not training time; training cost is therefore not included in CR, and encoding/decoding runtime optimization is left for future work.

NGLR supports practical parallel execution. Once the base reconstruction is available, different blocks can be encoded and decoded independently. The convolutional bias predictor can also be evaluated in batched form on GPUs. Within each block, the residual recovery follows the fixed causal scan order required by the \(q\)-neighbor context.

\section{Experiments}
\label{sec:experiments}

\subsection{Datasets}
We evaluate on three large-scale scientific datasets that span climate, turbulence, and atmospheric reanalysis. Table~\ref{tab:datasets} summarizes their dimensions. All values are stored in float32.

\begin{table}[h]
\centering
\caption{Evaluation datasets. The five shape dimensions correspond to variables, regions, time steps, and the two spatial dimensions.}
\label{tab:datasets}
\setlength{\tabcolsep}{4pt}
\small
\begin{tabular}{llcc}
\toprule
\textbf{Dataset} & \textbf{Domain} & \textbf{Shape} & \textbf{Size} \\
\midrule
E3SM~\cite{e3sm}   & Climate            & $1\!\times\!6\!\times\!720\!\times\!240\!\times\!240$  & 1 GB \\
ERA5~\cite{era5}   & Atm.\ reanalysis   & $1\!\times\!1\!\times\!960\!\times\!512\!\times\!512$  & 1 GB \\
JHTDB~\cite{jhtdb} & Fluid dynamics     & $1\!\times\!4\!\times\!240\!\times\!512\!\times\!512$  & 1 GB \\
\bottomrule
\end{tabular}
 \vspace*{-0.3cm}
\end{table}

These datasets span different residual-predictability regimes. E3SM~\cite{e3sm} is a high-resolution Earth system simulation with complex spatiotemporal structure. ERA5~\cite{era5} is the European Centre for Medium-Range Weather Forecasts (ECMWF) global atmospheric reanalysis, providing smoother large-scale climate fields. JHTDB~\cite{jhtdb} provides direct numerical simulation of turbulent flow with fine-scale structure that is particularly difficult to compress at tight error targets.

\subsection{Baselines}
We compare four methods at each NRMSE target:
\begin{itemize}
\item \textbf{SZ}~\cite{SZ_3,sz3}: A strong traditional error-bounded compressor based on Lorenzo prediction and entropy coding.
\item \textbf{GAE}~\cite{jaemoon3}: The Guaranteed Autoencoder framework, which uses a global PCA-style projection onto the SVD basis of the decoder linear operator for residual correction.
\item \textbf{LBRC} (ours, Section~\ref{sec:lbrc}): The training-free 3D Lorenzo residual coder applied after the same base learned compressor used by GAE.
\item \textbf{NGLR} (ours, Section~\ref{sec:nglr}): The neural-guided Lorenzo residual coder, also applied to the same base learned compressor.
\end{itemize}

For both LBRC and NGLR, we use a CAESAR-V~\cite{caesar} variational autoencoder with hyperpriors as the base learned compressor \(\mathcal{C}\), configured identically to the base model in the GAE pipeline so that all three learned methods share the same \(\tilde{x}\) and differ only in the residual correction strategy.
We use SZ3 (\texttt{v3.2.1-38-gdd09ba7}) through the \texttt{pysz} interface in pointwise absolute-error mode on the normalized field. For each target, the SZ absolute tolerance is selected by one-dimensional search as the largest tolerance whose decoded output satisfies the block-level NRMSE target.
Because pointwise tolerance and block-level NRMSE are related but not equivalent, all rate-distortion curves are plotted using the achieved decoded block-level NRMSE rather than nominal tolerance values. This ensures that compression ratios across methods are compared using the achieved reconstruction accuracy.
\begin{figure*}[t]
\centering
\subfloat[E3SM (climate)]{
    \includegraphics[width=0.30\textwidth]{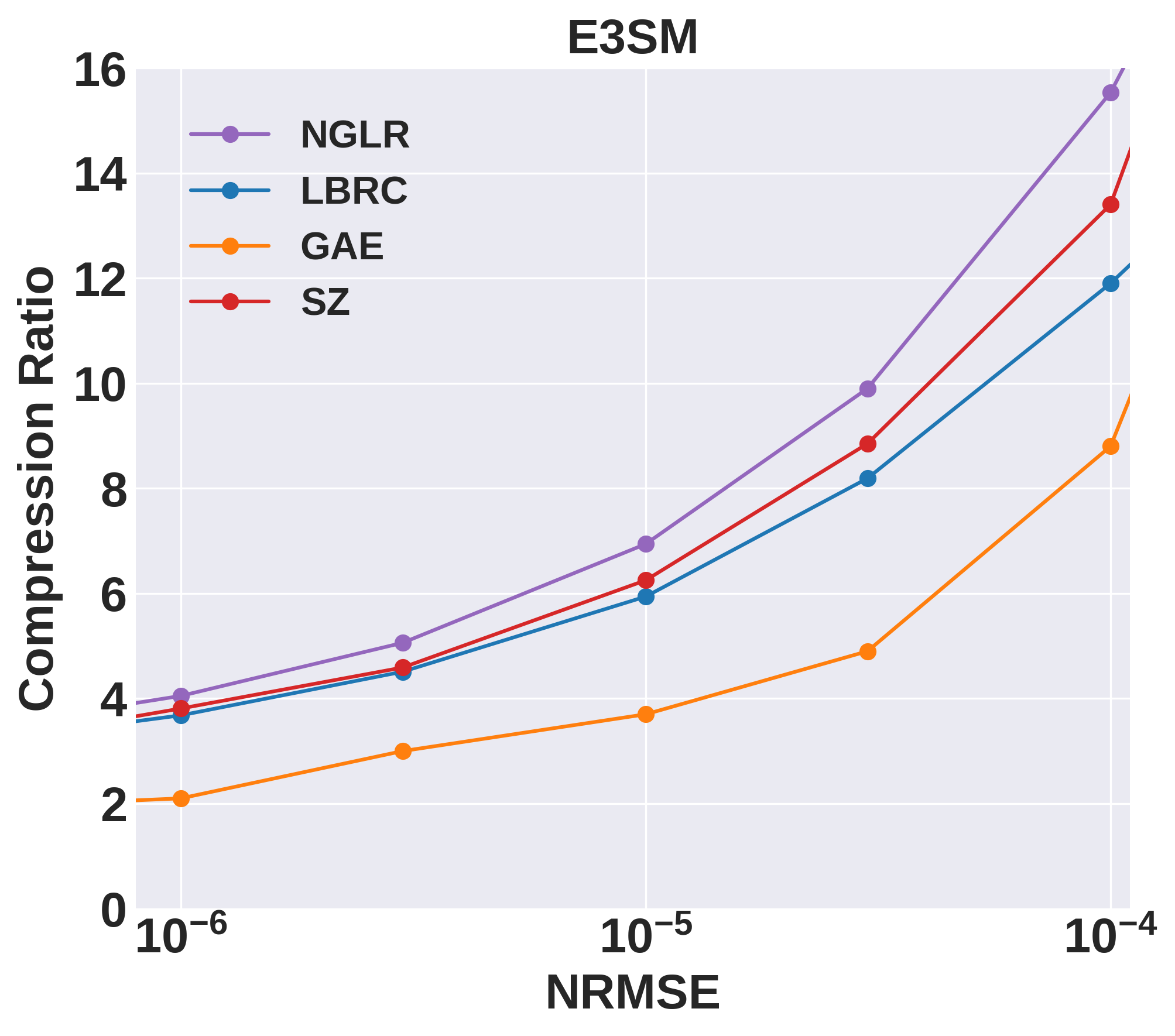}
    \label{fig:e3sm}
}
\hfill
\subfloat[JHTDB (turbulence)]{
    \includegraphics[width=0.30\textwidth]{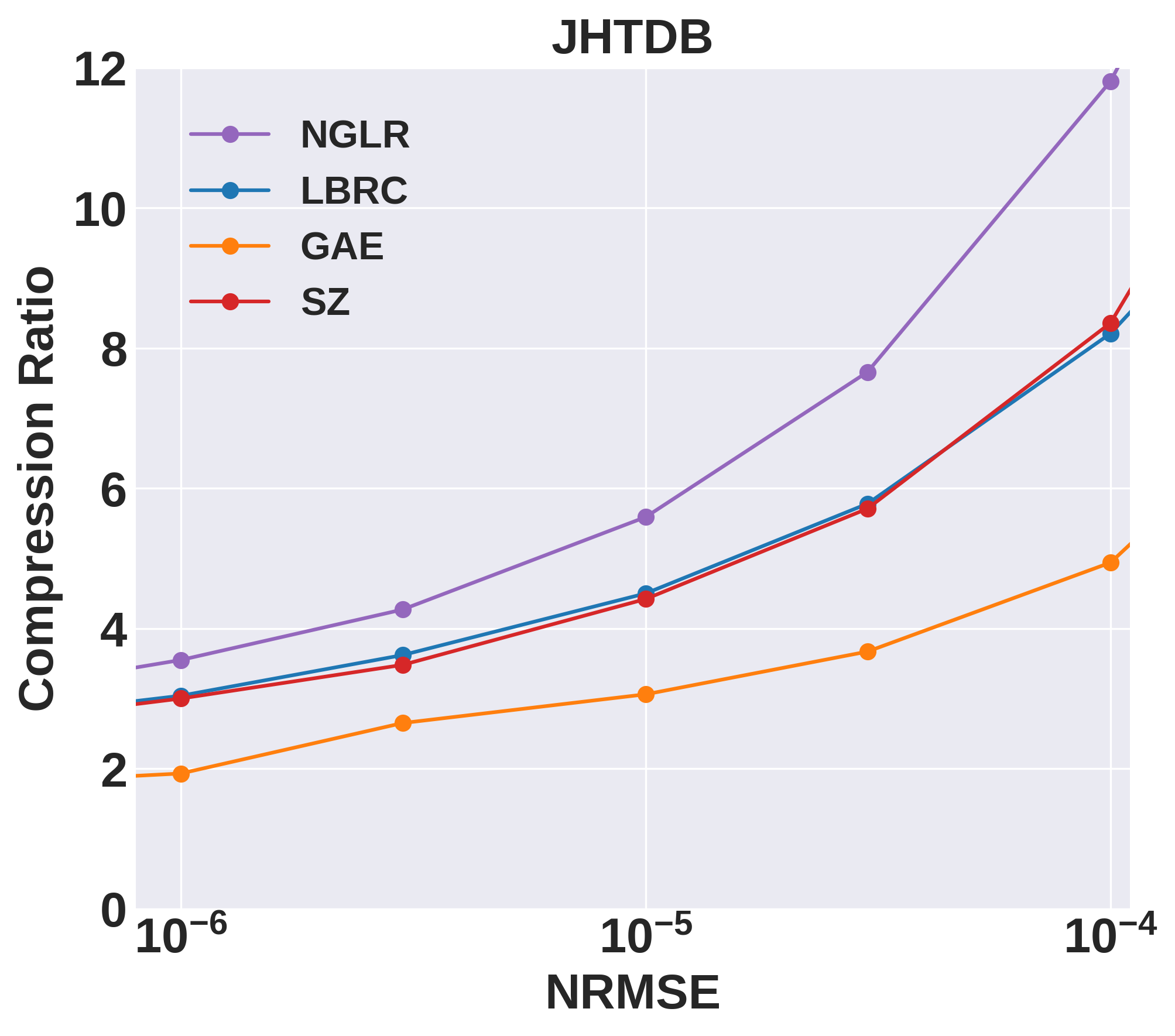}
    \label{fig:jhtdb}
}
\hfill
\subfloat[ERA5 (atmospheric reanalysis)]{
    \includegraphics[width=0.30\textwidth]{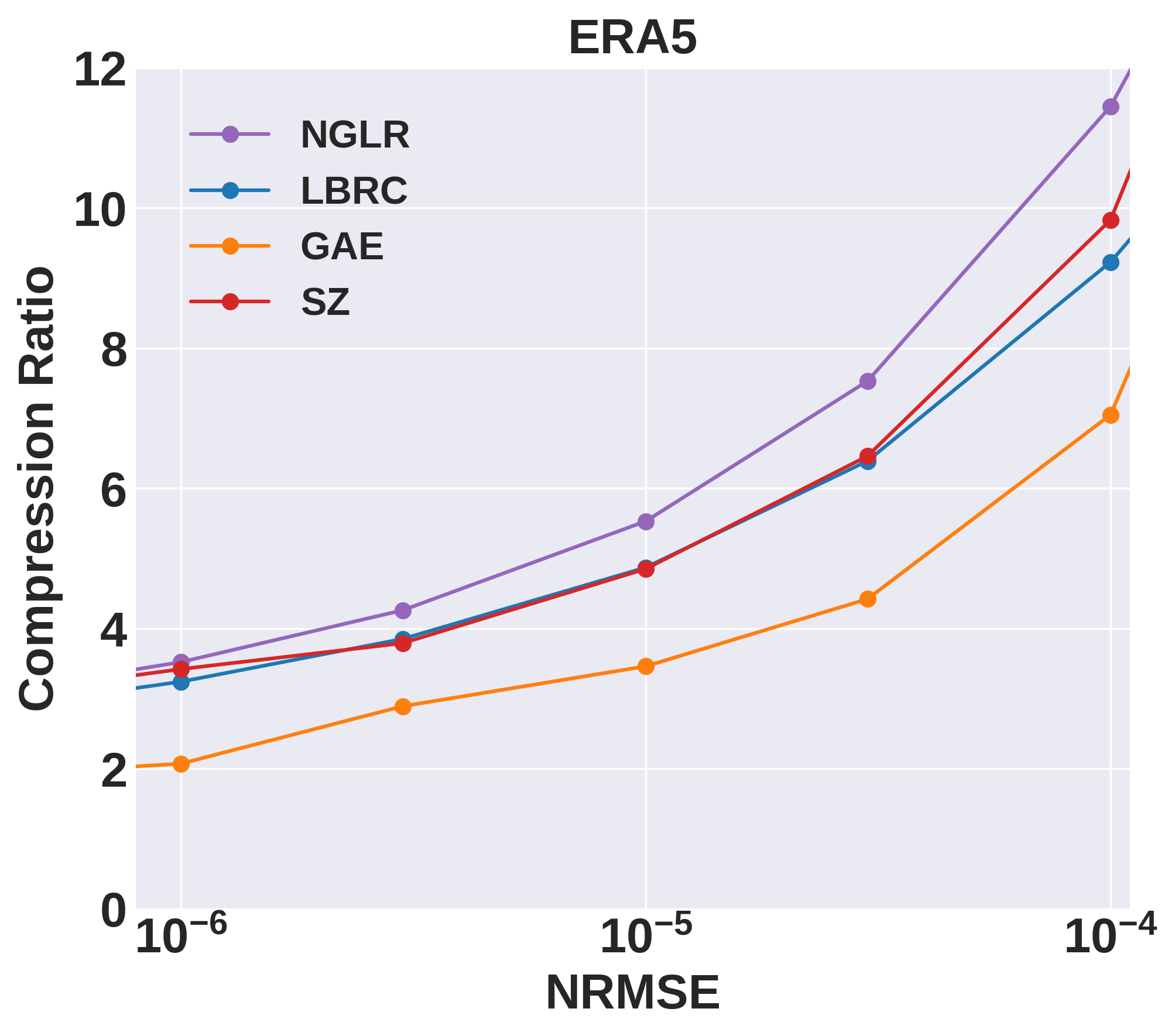}
    \label{fig:era5}
}
\caption{Compression ratio vs.\ achieved block-level NRMSE across the three datasets. Across datasets, NGLR gives the highest CR, GAE the lowest, and LBRC is broadly competitive with SZ. The NGLR--GAE gap widens at tighter NRMSE, indicating that GAE's per-block coefficient cost is what erodes the learned-base advantage at high fidelity. The size of the NGLR--LBRC gap varies by dataset: it is largest on JHTDB, where fine-scale turbulent residual structure is hardest to capture with the fixed Lorenzo stencil and benefits most from neural guidance; it is smallest on ERA5 at $10^{-6}$, where the learned residual approaches white noise and local prediction has less structure to exploit, an intrinsic limit of any local predictor on a decorrelated residual.}
\label{fig:results}
\vspace*{-0.6cm}
\end{figure*}

\subsection{Implementation Details}

All experiments were run on the HiPerGator computing system with NVIDIA B200 180\,GB GPUs. For all learned-residual methods, we use the same base reconstruction \(\tilde{x}\) produced by CAESAR-V~\cite{caesar}. For NGLR, the neural bias predictor is trained for each compressed collection and target accuracy. Its serialized weights are stored as part of the compressed representation and counted once in the total compressed size, rather than treated as a free external model.

The neural bias predictor follows the architecture in Fig.~\ref{fig:nglr}: a Guide Encoder with a Conv3D layer plus four 3D residual blocks, a Context Encoder with two \(1\times1\times1\) Conv3D layers, and a per-voxel Fusion module. The causal context input contains 8 channels: 7 first-order Lorenzo \(q\)-neighbors and the corresponding standard Lorenzo estimate. Unless otherwise specified, we use 32 Guide hidden channels and 16 context hidden channels. This configuration has approximately \(2.3\times10^5\) trainable parameters, and the serialized model size is about \(0.94\) MB. For each dataset and target, the predictor is trained on blocks from the same compressed collection using AdamW with learning rate $5\times 10^{-4}$ for 50 epochs. In each epoch, all blocks are randomly shuffled, and the model is updated once per block. The resulting weights are serialized once and included in the compressed size.

We sweep the NRMSE target $\tau$ over $\{10^{-6}, 3\times10^{-6}, 10^{-5}, 3\times10^{-5}, 10^{-4}\}$ and report the achieved compression ratio, defined as the ratio of the original float32 size to the total compressed size including the shared CAESAR-V base latent stream, the residual stream (and, for NGLR, the amortized bias-predictor weight cost). Throughout this section, ``NRMSE'' values on the x-axis of every figure, and in the per-dataset paragraphs below, denote \emph{achieved} block-level NRMSE on the evaluated blocks; this ensures that CR comparisons across methods are always at matched (or no-worse) reconstruction accuracy.

\subsection{Results on E3SM}
\label{sec:e3sm-results}
Figure~\ref{fig:results}(a) shows the compression ratio as a function of NRMSE on E3SM. In this high-fidelity regime, GAE achieves the lowest CR across all error targets: the global PCA-style residual correction becomes costly as the target accuracy tightens, because the number of basis coefficients needed to reach the bound grows rapidly. LBRC consistently improves over GAE, with the gap widening at tighter targets where the global correction is most expensive; the local Lorenzo residual coding is far more efficient than storing global residual coefficients in this regime.

NGLR further improves on LBRC by adding neural guidance, giving roughly 15\% or more CR gain over LBRC at every target. Compared with SZ, NGLR is also consistently better, with about 10\% average improvement. For example, at NRMSE $=10^{-4}$, NGLR reaches CR $=15.5$ versus 11.9 for LBRC, 13.4 for SZ, and 8.8 for GAE; the ordering and the size of the gap are preserved at $10^{-5}$ and $10^{-6}$. Overall, neural-guided Lorenzo coding preserves the advantage of the base learned compressor in the high-accuracy regime where a GAE-style correction would otherwise erode it.

\subsection{Results on JHTDB}
\label{sec:jhtdb-results}
Figure~\ref{fig:results}(b) shows the results on JHTDB, where the data exhibits fine-scale turbulent structure that is particularly hard to compress at tight error targets.

In this high-fidelity regime, GAE again performs the worst, confirming that global PCA-style residual correction is inefficient at these low-NRMSE targets. LBRC brings a clear improvement over GAE, with roughly 50--60\% average CR gain. SZ and LBRC are close on this dataset, which indicates that the training-free Lorenzo residual coder is already competitive with a strong traditional compressor in this regime. The key result, however, is that NGLR clearly separates from both LBRC and SZ after adding neural guidance. Across the tested NRMSE range, NGLR improves over LBRC by roughly 15--45\% and over SZ by 20--40\%. At $10^{-4}$, NGLR reaches CR $=11.8$ versus 8.21 for LBRC and 8.36 for SZ. Overall, JHTDB provides evidence that NGLR not only matches traditional compressors but can clearly outperform them when residual structure can be learned and summarized by the neural bias model.

\subsection{Results on ERA5}
\label{sec:era5-results}
Figure~\ref{fig:results}(c) shows the results on ERA5, an atmospheric reanalysis dataset with smoother variation than JHTDB but more variability than E3SM.
As on the other two datasets, GAE has the lowest CR at these high-fidelity targets. LBRC improves over GAE by roughly 30--55\%. SZ is particularly competitive at the strictest target ($10^{-6}$), where the learned residual approaches white noise and the local-prediction advantage narrows---an intrinsic limit of any local predictor on a decorrelated residual. NGLR nonetheless maintains the best overall performance: it improves over LBRC by about 10--25\% and exceeds SZ by roughly 10\% on average. At $10^{-4}$, NGLR achieves CR $=11.45$ versus 9.23 for LBRC, 9.83 for SZ, and 7.05 for GAE. Across E3SM, JHTDB, and ERA5, the trend is consistent: neural-guided Lorenzo residual coding reduces the high-fidelity correction cost and preserves the advantage of the learned base reconstruction.

\subsection{Additional Analyses}
\label{sec:planned}

\subsubsection{NGLR ablation}

Table~\ref{tab:nglr_ablation} ablates the two information sources in NGLR on JHTDB. Both single-branch variants improve over LBRC, confirming that base-reconstruction features and causal q-neighbor context each provide useful predictive information. The full model gives the highest CR at all three NRMSE targets, indicating that the two branches are complementary: reconstruction features provide continuous-field structural context, while q-neighbors provide causal code-domain context in the quantized residual. The gain is largest at $10^{-4}$ and narrows at $10^{-6}$, consistent with reduced residual predictability at stricter tolerances.

Since GAE, LBRC, and NGLR share the same CAESAR-V base reconstruction, their base-latent cost is identical; the CR differences in Fig. 3 therefore reflect differences in the residual correction stream, with NGLR’s serialized predictor weights included in the total size. The improvement of LBRC over GAE shows that local predictive coding is more compact than global per-block PCA correction in the high-fidelity regime. The additional improvement of NGLR shows that the learned bias predictor further reduces the remaining  residual code.

\begin{table}[t]
\centering
\caption{
Ablation of NGLR guidance components on JHTDB. NGLR-Recons uses only reconstruction guidance, while NGLR-QNeighbor uses only causal q-neighbor guidance. Both improve over LBRC, and full NGLR performs best, showing the two guidance sources are complementary.
    }
\label{tab:nglr_ablation}
\setlength{\tabcolsep}{3pt}
\renewcommand{\arraystretch}{1.15}
\begin{tabular}{c|cccc}
\textbf{Achieved NRMSE} & \multicolumn{4}{c}{\textbf{Compression Ratio (CR)}} \\
\cline{2-5}
 & \textbf{NGLR} & \textbf{NGLR-Recons} & \textbf{NGLR-QNeighbor} & \textbf{LBRC} \\
\hline
$1\times10^{-4}$ & 11.81 & 9.41 & 9.04 & 8.21 \\
$1\times10^{-5}$ & 5.60  & 5.03 & 4.77 & 4.50 \\
$1\times10^{-6}$ & 3.55  & 3.31 & 3.17 & 3.04 \\
\hline
\end{tabular}
 \vspace*{-0.6cm}
\end{table}

\subsubsection{Runtime}

Table~\ref{tab:runtime} reports an initial throughput test on the JHTDB dataset at the NRMSE target $10^{-5}$ using an NVIDIA B200 GPU. The reported values isolate the residual-correction stage; the shared CAESAR-V base compression cost is the same across all learned methods and is not included in Table~\ref{tab:runtime}. The NGLR compression throughput includes predictor inference, residual coding, and entropy coding; it excludes predictor training and base-compressor cost. NGLR predictor training is performed separately and takes about 20--30 minutes on a 1\,GB collection. Because CR measures stored bits rather than wall-clock cost, training time is reported separately and not included in CR. For NGLR, the intra-block causal dependency is handled by a causal diagonal scan, which processes voxels on the same diagonal together while constructing each causal $q$-context only from previously decoded values.
NGLR is slower than LBRC and GAE because it adds neural bias prediction and an intra-block causal scan, but it also gives the highest CR at this accuracy target: 114.2 / 157.0 MB/s for NGLR, versus 145.1 / 1138.9 MB/s for LBRC and 412.7 / 837.0 MB/s for GAE (compress / decompress). LBRC therefore remains the low-overhead option, while NGLR is usable for settings where improved compression ratio is prioritized over peak decompression throughput. 

\subsection{Discussion}
The experiments confirm three findings. The results are consistent with GAE-style global correction dominating the rate cost in the regime $\tau \in [10^{-6}, 10^{-4}]$, matching the bottleneck noted for CAESAR; LBRC, a local prediction-based coder, already recovers most of the lost advantage without learned components; and a lightweight neural bias predictor (NGLR) adds further improvement while preserving deterministic decoding. The dataset variation also shows the limit of the approach: NGLR provides the largest gains when learned residuals retain structured local patterns, while gains narrow when the residual approaches decorrelated noise.

A practical advantage of both LBRC and NGLR is accurate target matching. Because the residual quantization step is selected by binary search, the achieved NRMSE closely matches the requested $\tau$ rather than being far below it (which would unnecessarily sacrifice compression ratio). This makes the methods promising for HPC pipelines where compression budgets are set by downstream accuracy requirements.

A key aspect of the proposed framework is the separation of fidelity control from learned entropy reduction. Adaptive residual quantization establishes the desired block-level NRMSE guarantee \textit{before} any neural processing is applied. The subsequent predictive and neural stages operate entirely on an integer representation whose reconstruction is fixed, allowing learned models to improve compression efficiency without affecting correctness or error guarantees. 

Together, LBRC and NGLR form a practical spectrum: deployments can pick the operating point matching the encode/decode compute budget.

\begin{table}[t]
\centering
\caption{Compression and decompression throughput of residual correction methods.}
\label{tab:runtime}
\setlength{\tabcolsep}{3pt}
\renewcommand{\arraystretch}{1.15}
\begin{tabular}{lcc}
\hline
Method & Compress (MB/s) & Decompress (MB/s) \\
\hline
NGLR & 114.20 & 157.04 \\
LBRC & 145.12 & 1138.93 \\
GAE  & 412.73 & 837.03 \\
\hline
\end{tabular}
 \vspace*{-0.6cm}
\end{table}

\section{Conclusion}
\label{sec:conclusion}
We presented LBRC and NGLR, two residual coders for high-fidelity learned scientific compression. LBRC replaces global PCA-style correction with target-matched integer Lorenzo residual coding, while NGLR adds a causal neural bias predictor whose weights are charged to the compressed stream. Together, they show that the residual representation becomes the rate-limiting design choice once the accuracy target is tight enough for correction bits to dominate the rate. These results support the view that the learned residual should be treated as a coding object in its own right.

Across E3SM, JHTDB, and ERA5 at NRMSE targets between $10^{-6}$ and $10^{-4}$, LBRC improves CR over GAE by 30--60\%, and NGLR adds a further 10--40\% over LBRC. The takeaway is that the residual representation, not only the base learned model, drives high-fidelity learned scientific compression. Future work includes adaptive context shapes that vary with local field structure, integration into open-source scientific compression frameworks, and evaluation on additional domains such as combustion and astrophysics.



\section*{Acknowledgment}
OpenAI ChatGPT was used solely for copy-editing. All technical content remains the authors’ responsibility. The work of Anand Rangarajan and Sanjay Ranka was supported in part by UT-Battelle, LLC, under contract DE-AC05-00OR22725 with the U.S. Department of Energy (DOE).

\end{document}